\documentclass[journal]{IEEEtai}

\usepackage[colorlinks,urlcolor=blue,linkcolor=blue,citecolor=blue]{hyperref}

\usepackage{color,array}

\usepackage{graphicx}

\usepackage[utf8]{inputenc}
\usepackage{booktabs} 
\usepackage{multirow} 
\usepackage{graphicx}
\usepackage[table,xcdraw]{xcolor}
\usepackage{algorithm}
\usepackage{algpseudocode}
\usepackage{amsmath}
\usepackage{amssymb}

\begin{document}

\title{Efficient and Interpretable Multi-Agent LLM Routing via Ant Colony Optimization} 

\author{Xudong Wang, Chaoning Zhang*,~\IEEEmembership{Senior Member,~IEEE,} Jiaquan Zhang, Chenghao Li, Qigan Sun, \\
Sung-Ho Bae,~\IEEEmembership{Member,~IEEE,} Peng Wang, Ning Xie, Jie Zou, \\
Yang Yang,~\IEEEmembership{Senior Member,~IEEE,} and Hengtao Shen,~\IEEEmembership{Fellow,~IEEE}

\thanks{Xudong Wang, Qigan Sun, and Sung-Ho Bae are with the School of Computing, Kyung Hee University, Yongin-si 17104, South Korea (e-mail: wl200203@khu.ac.kr; sunqigan0206@gmail.com; shbae@khu.ac.kr). Jiaquan Zhang is with the School of Information and Software Engineering, University of Electronic Science and Technology of China, Chengdu 610054, China (e-mail: jiaquanzhang2005@gmail.com). Chaoning Zhang, Chenghao Li, Peng Wang, Ning Xie, Jie Zou, and Yang Yang are with the School of Computer Science and Engineering, University of Electronic Science and Technology of China, Chengdu 611731, China (e-mail: chaoningzhang1990@gmail.com; lch17692405449@gmail.com; wangpeng8619@gmail.com; xiening@uestc.edu.cn; jie.zou@uestc.edu.cn; yang.yang@uestc.edu.cn). Hengtao Shen is with the School of Computer Science and Technology, Tongji University, Shanghai 200092, China (e-mail: shenhengtao@hotmail.com).}
\thanks{* Corresponding Author}
}

\markboth{Journal of IEEE Transactions on Artificial Intelligence, Vol. 00, No. 0, Month 2020}
{First A. Author \MakeLowercase{\textit{et al.}}: Bare Demo of IEEEtai.cls for IEEE Journals of IEEE Transactions on Artificial Intelligence}

\maketitle

\begin{abstract}
Large Language Model (LLM)-driven Multi-Agent Systems (MAS) have demonstrated strong capability in complex reasoning and tool use, and heterogeneous agent pools further broaden the quality--cost trade-off space. Despite these advances, real-world deployment is often constrained by high inference cost, latency, and limited transparency, which hinders scalable and efficient routing. Existing routing strategies typically rely on expensive LLM-based selectors or static policies, and offer limited controllability for semantic-aware routing under dynamic loads and mixed intents, often resulting in unstable performance and inefficient resource utilization. To address these limitations, we propose AMRO-S, an efficient and interpretable routing framework for Multi-Agent Systems (MAS). AMRO-S models MAS routing as a semantic-conditioned path selection problem, enhancing routing performance through three key mechanisms: First, it leverages a supervised fine-tuned (SFT) small language model for intent inference, providing a low-overhead semantic interface for each query; second, it decomposes routing memory into task-specific pheromone specialists, reducing cross-task interference and optimizing path selection under mixed workloads; finally, it employs a quality-gated asynchronous update mechanism to decouple inference from learning, optimizing routing without increasing latency. Extensive experiments on five public benchmarks and high-concurrency stress tests demonstrate that AMRO-S consistently improves the quality--cost trade-off over strong routing baselines, while providing traceable routing evidence through structured pheromone patterns.
\end{abstract}

\begin{IEEEImpStatement}
Large language model (LLM)-based multi-agent systems can improve automated problem solving, but practical deployment is often limited by cost, latency, and weak transparency, especially under high concurrency. This paper introduces AMRO-S, which combines small language models with ant colony optimization for efficient and interpretable routing in multi-agent systems. AMRO-S delivers up to 4.7$\times$ speedup, reduces inference cost, and maintains strong accuracy across diverse benchmarks. It also provides semantically meaningful routing evidence through pheromone specialists, supporting diagnosis and trust in latency- and resource-constrained settings, including high-stakes applications.
\end{IEEEImpStatement}


\section{Introduction}

\IEEEPARstart{L}{arge} Language Models (LLMs) have achieved substantial progress in natural language understanding, multi-step reasoning, and code generation, catalyzing the rapid development of LLM-driven agents and Multi-Agent Systems (MAS). MAS can be viewed as distributed systems composed of multiple LLM-based agents that communicate, collaborate, and coordinate to accomplish complex tasks \cite{li2025parallelized,zhuge2024gptswarm,mosquera2025can}. By decomposing complex problems into subtasks and leveraging heterogeneous agents with distinct capability--cost profiles, MAS demonstrate strong scalability and performance in domains such as automated programming \cite{yang2025docagent,yuan2024evoagent}, mathematical reasoning \cite{motwani2024malt}, and collaborative decision-making \cite{jin2025comprehensive,wang2024cooperation}.

However, as MAS grow in scale and task distributions become increasingly diverse, MAS routing has emerged as a key bottleneck in dynamic and resource-constrained environments \cite{li2024survey,aratchige2025llms,chacon2025cooperative}. For each incoming request, the system must select an appropriate execution path from a heterogeneous agent pool while jointly balancing output quality and serving overhead, including latency, token usage, and load. In many engineering-oriented MAS frameworks, routing is still implemented via two relatively simplified paradigms: static rule-based allocation, which relies on predefined templates or fixed topologies and thus adapts poorly to load fluctuations and node availability changes, and full-context broadcasting to all agents, which offers implementation simplicity but incurs significant redundancy in tokens and computation \cite{liu2025rcr}. Under high-concurrency and low-latency constraints, these strategies often lead to limited throughput, degraded latency, and escalating costs \cite{cemri2025multi}. While recent advances attempt to mitigate such information redundancy by extracting global structural invariants via topological data analysis \cite{zhang2026text}, applying this structural awareness directly to dynamic MAS routing introduces non-trivial computational overhead. This naturally leads to the following question: \emph{How can we balance quality, cost, and latency in semantic-aware, path-level routing under time-varying system conditions and mixed user intents?}

\begin{figure}[htbp]
    \centering
    \includegraphics[width=1\linewidth]{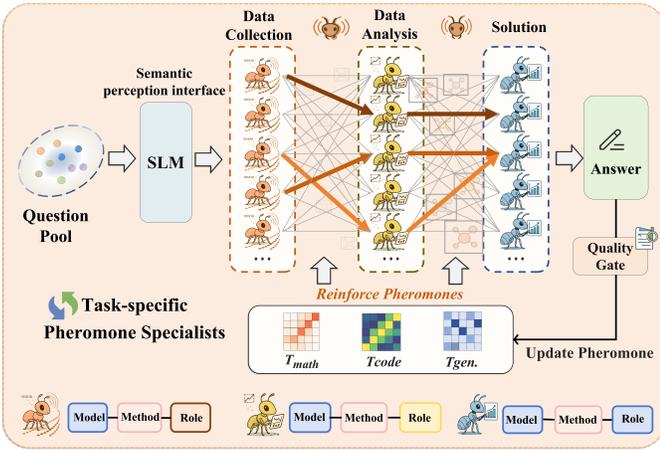}
    \caption{Overview of the AMRO-S routing mechanism. Tasks are routed through three stages, collection, analysis, and solution, via probabilistic path sampling guided by dynamic pheromone signals. After execution, high-quality paths receive reinforced pheromones, increasing their selection likelihood.}
    \label{fig:1}
\end{figure}

Recent studies have explored LLM selection and dynamic routing by using LLMs to match task semantics and route queries to suitable agents \cite{chuang2025confident,wang2024survey,yue2025masrouter,zhao2026tcandon}. Related efforts also emphasize multi-stage collaboration through hierarchical workflows, knowledge structures, and role allocation to better handle complex tasks \cite{wang2024mixture,zhang2025agentrouter,liu2025rcr,hu2024scalable}. Despite this progress, realistic MAS deployments still face recurring challenges. Routing decisions are often buried in black-box inference or opaque selectors, which limits transparency in high-stakes domains such as healthcare and finance \cite{marey2024explainability,chacon2025cooperative}. Many routing policies remain static or semi-static and respond poorly to changes in node load, network fluctuations, and task dynamics, leading to unstable performance under mixed workloads. In addition, deployment cost remains non-trivial, since some approaches rely on large-scale annotations or expensive training procedures that are difficult to justify in edge computing or strict low-latency scenarios \cite{varangot2025doing,zhang2024edgeshard}. Although reward-based and meta-learning strategies have been explored to improve adaptability, they often introduce complex designs and high training overhead, hindering widespread adoption. Overall, a unified routing mechanism that couples semantic modeling, task-isolated memory, and controllable online updates under strict serving constraints remains underexplored.

To address these limitations, we propose AMRO-S, an efficient and interpretable routing framework for heterogeneous MAS. AMRO-S models MAS routing as semantic-conditioned path selection on a layered directed graph, as illustrated in Fig.~\ref{fig:1}. The framework is supported by three synergistic mechanisms. First, it leverages a supervised fine-tuned (SFT) small language model for intent inference, providing a low-overhead semantic interface for each query. Second, inspired by the biological logic of path search guided by pheromones in Ant Colony Optimization (ACO) \cite{dorigo2018ant}, routing memory is factorized into task-specific pheromone specialists, and query-conditioned fusion is applied to reduce cross-task interference and optimize path selection under mixed workloads. Finally, AMRO-S employs a quality-gated asynchronous update mechanism to decouple inference from learning, reinforcing pheromone specialists only with high-quality trajectories in the background, refining routing without increasing serving latency.

We evaluate AMRO-S on diverse reasoning and coding benchmarks and under high-concurrency stress tests. Results show that AMRO-S improves the average score by 1.90 points over the strongest multi-agent routing baseline and achieves up to 4.7$\times$ speedup under 1000 concurrent processes, while maintaining stable latency under load. In addition, structured pheromone patterns provide traceable routing evidence, enabling transparent diagnosis and continual optimization.

Our contributions are summarized as follows:
\begin{itemize}
    \item We introduce AMRO-S, which models MAS routing as semantic-conditioned path selection on a layered directed graph with explicit quality--cost considerations.
    \item We propose task-specific pheromone specialists with query-conditioned fusion to isolate task memories and mitigate cross-task interference under mixed intents.
    \item We develop a quality-gated asynchronous update mechanism for controllable online optimization under strict serving constraints.
    \item We demonstrate improvements in accuracy, cost-efficiency, and stability across benchmarks and provide path-level interpretability via pheromone pattern analyses.
\end{itemize}

\section{Related Work}
\subsection{LLM-based Multi-Agent System Routing}
MAS are composed of multiple agents with autonomous perception, learning, and decision-making capabilities, enabling them to complete complex tasks through distributed collaboration \cite{dorri2018multi}. They overcome the limitations of single-agent systems in memory capacity and scalability \cite{balaji2010introduction}. LLM-based MAS integrate the powerful language understanding capabilities of LLMs \cite{kasneci2023chatgpt,zheng2025towards,chacon2025cooperative} with group-level strategy coordination abilities \cite{li2024survey,han2024llm,mosquera2025can}, further enhancing their problem-solving capacity for complex tasks. To improve system efficiency, LLM routing precisely allocates user requests to appropriate subagents, tools, plugins, or modules based on task content \cite{hu2024routerbench}, making the design of effective routing strategies a current research focus. AGENTVERSE \cite{chen2023agentverse} dynamically determines the composition of the agent through an expert recruitment phase. MAD \cite{liang2023encouraging} designs a multi-agent debate structure with sparse communication topologies, achieving comparable performance while significantly reducing computational costs \cite{liang2023encouraging,adornetto2025generative}. Similarly, recent advances leverage topological structural modeling to extract non-redundant reasoning chains among diverse agents \cite{zhanglearning}. However, non-learnable path strategies in complex tasks restrict model generalization and flexibility. ZOOTER \cite{lu2023routing} proposes reward-guided routing, extracting rewards from training queries to train a routing function that assigns each query to an LLM with relevant expertise. RouterDC \cite{chen2024routerdc} learns a query-based router using sample-LLM and inter-sample contrastive loss functions. Hybrid-LLM \cite{ding2024hybrid} introduces a hybrid LLM routing method to improve reasoning efficiency by combining the advantages of multiple LLMs. RouteLLM \cite{ong2024routellm} optimizes the balance between cost and response quality through dynamic selection of strong and weak LLMs, while MasRouter \cite{yue2025masrouter} addresses complex routing problems using a three-level cascaded framework for collaboration mode determination, role allocation, and routing assignment.

\subsection{Heuristic path optimization}
Heuristic path optimization rapidly searches for optimal or near-optimal paths through empirical strategies \cite{tan2021comprehensive,yahia2023path,hu2024scalable}. Classic heuristic path optimization algorithms include genetic optimization algorithms \cite{sivanandam2008genetic}, simulated annealing algorithms \cite{rutenbar1989simulated}, and particle swarm optimization algorithms \cite{wang2018particle,gad2022particle1}, among others \cite{li2025distributed}. The ant colony algorithm, in particular, provides effective optimization strategies for fields such as path planning \cite{cui2024multi}, network routing, and vehicle scheduling due to its feedback mechanism and strong parallel computing characteristics. ACO algorithm is an optimization method inspired by the foraging behavior of natural ant colonies \cite{blum2005ant,dorigo2018introduction}. In nature, ants indirectly communicate by releasing pheromones while searching for food. Other ants prefer paths with higher pheromone concentrations, as these typically indicate better routes. This mechanism forms a positive feedback loop, guiding more ants to follow optimal paths until the colony identifies the shortest route from the nest to food sources. AddACO \cite{scianna2024addaco} proposed incorporating decision rules based on linear convex combinations into the ant colony algorithm, improving the computational efficiency for the Traveling Salesman Problem (TSP). DYACO \cite{liang2024enhanced} optimized the impact of complex slopes in deep-sea mining areas on path planning by dynamically adjusting key information such as heuristic guidance, significantly enhancing the convergence speed of path optimization. PACO \cite{si2024novel} addressed the local optimum problem of traditional ACO through improved pheromone update methods and hybrid strategies, and substantially boosted path planning efficiency via parallel computing.

Despite the good performance achieved by the above methods, practical applications still demand higher training efficiency and accuracy. Additionally, the black-box nature of LLMs limits the interpretability of routing. To address these issues, we introduce ACO and design a multi-agent routing mechanism, enabling the MAS to maintain low cost, high efficiency, and high concurrent processing capabilities while enhancing interpretability.

\section{Methodology}
\label{sec:methodology}

\begin{figure*}[t]
    \centering
    \includegraphics[width=\linewidth]{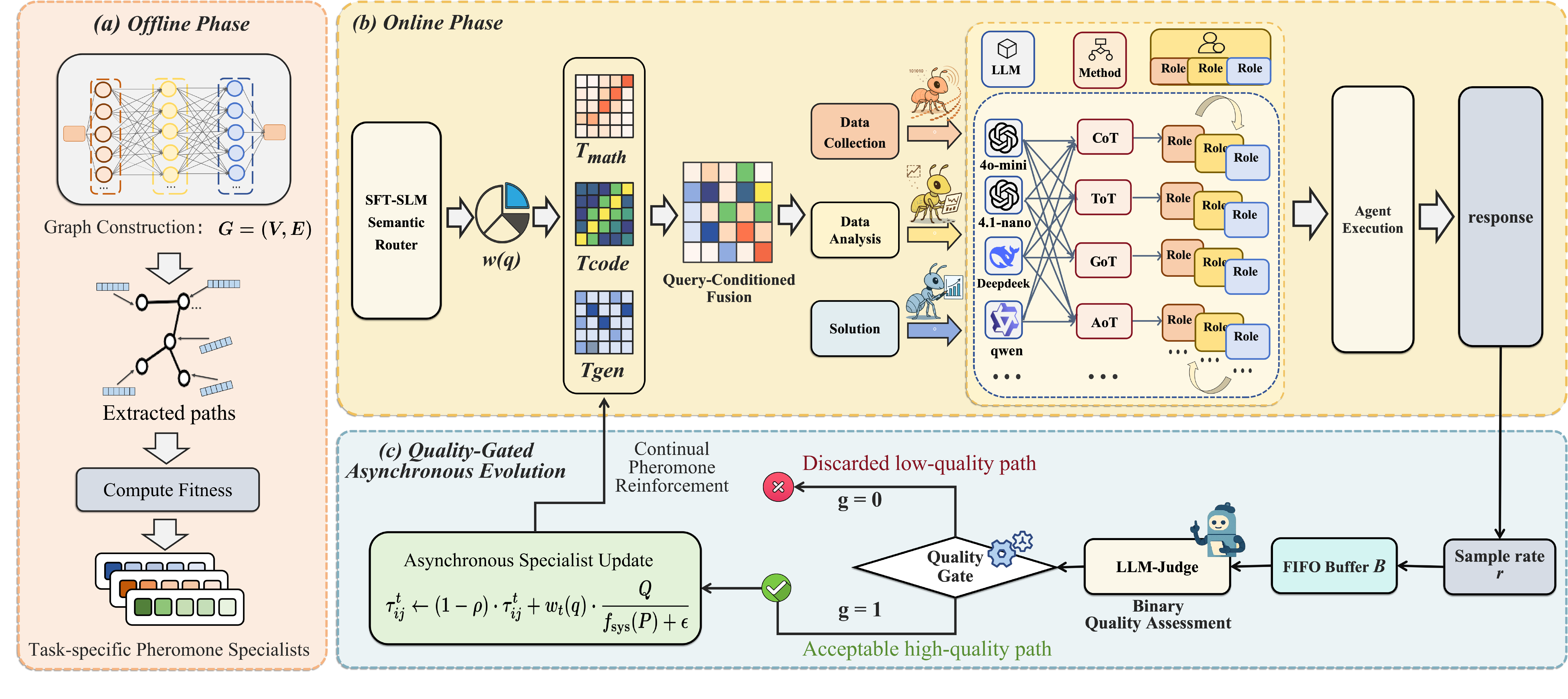}
    \caption{Architecture of AMRO-S. (a) Offline construction of layered graph $G=(V,E)$ and pheromone specialists. (b) Online routing via SFT-SLM weights $w(q)$ across three stages, where nodes represent (LLM, Method, Role) instances. (c) Asynchronous evolution using LLM-Judge quality gating ($g \in \{0,1\}$) for background pheromone reinforcement without serving overhead.}
    \label{fig:amro_arch}
\end{figure*}

This section presents \textbf{AMRO-S}, a routing framework for multi-agent systems (MAS) under (i) heterogeneous agent capabilities, (ii) mixed user intents, and (iii) time-varying system load. We first establish the problem formulation and graph modeling in Section~\ref{subsec:graph_model}. Subsequently, as illustrated in Figure~\ref{fig:amro_arch}, AMRO-S consists of three core components: 
(1) an SLM-based semantic task router, detailed in Section~\ref{subsec:slm_router}; 
(2) task-specific pheromone experts with query-conditioned fusion, as described in Section~\ref{subsec:pheromone_fusion}; and 
(3) online \emph{quality-gated asynchronous evolution} for continual adaptation without increasing serving latency, presented in Section~\ref{subsec:qgae}.

\subsection{Problem Formulation and Graph Modeling}
\label{subsec:graph_model}
We model routing in a multi-agent system (MAS) as a path-search problem on a layered directed graph $G=(V,E)$. The graph consists of $N$ processing stages (layers). Each layer $l\in\{1,\ldots,N\}$ contains $n$ heterogeneous agent nodes,
\begin{equation}
V_l=\{v_{l,1},v_{l,2},\ldots,v_{l,n}\}, \qquad
V=\bigcup_{l=1}^{N}V_l,
\end{equation}
where $v_{l,j}$ denotes the $j$-th agent instance at stage $l$. In our instantiation, each node corresponds to a fixed \emph{(backbone model $\times$ reasoning policy/role prompt)} pair, yielding diverse quality--cost profiles (e.g., different backbones combined with CoT/ToT/GoT/AoT or specialized role prompts). Directed edges exist only between adjacent layers,
\begin{equation}
E=\{(v_{l,i},v_{l+1,j}) \mid 1\le l < N,\ 1\le i,j\le n\},
\end{equation}
representing feasible transitions of the workflow from stage $l$ to stage $l+1$. For a given query $q$, a routing path is a node sequence from layer $1$ to layer $N$,
\begin{equation}
P=(v_{1,i_1},v_{2,i_2},\ldots,v_{N,i_N}).
\end{equation}
Our goal is to select an optimal path $P^*$ that balances task quality and system overhead:
\begin{equation}
P^{\ast} = \arg\max_{P} U(P;q), \qquad
U(P;q)=R(P;q)-\lambda\,C(P;q),
\label{eq:utility}
\end{equation}
where $R(P;q)$ measures task completion quality (e.g., answer correctness, unit-test pass rate, or judge-based quality signals), $C(P;q)$ is the system cost, and $\lambda>0$ controls the quality--cost trade-off.

To avoid ambiguous cost accounting, we explicitly decompose the cost into measurable components:
\begin{equation}
C(P;q)=\omega_{\mathrm{tok}}\cdot \mathrm{Tok}(P;q)
+\omega_{\mathrm{lat}}\cdot \mathrm{Lat}(P;q)
+\omega_{\mathrm{load}}\cdot \mathrm{Load}(P;q),
\label{eq:cost_decompose}
\end{equation}
where $\mathrm{Tok}(P;q)$ denotes token usage (or a monetary proxy converted from API pricing under a fixed accounting rule), $\mathrm{Lat}(P;q)$ is the end-to-end latency, and $\mathrm{Load}(P;q)$ aggregates node load statistics along the path (e.g., max or mean load; fixed across all methods in our setup). The weights $\omega_{\mathrm{tok}},\omega_{\mathrm{lat}},\omega_{\mathrm{load}}\ge 0$ align the objective with the inference-budget constraints and the unified cost accounting described in the experimental setup. Finally, to ensure executability and service stability under concurrency, we define the feasible candidate set for transitioning from node $v_{l,i}$:
\begin{equation}
\mathrm{Allowed}(l,i)=\{j \mid \mathrm{Avail}(v_{l+1,j})=1\ \wedge\ \mathrm{Load}(v_{l+1,j})\le \theta_{\mathrm{load}}\},
\label{eq:allowed}
\end{equation}
where $\mathrm{Avail}(\cdot)\in\{0,1\}$ indicates node availability (e.g., healthy endpoint, not circuit-broken), $\mathrm{Load}(\cdot)$ is the real-time load metric, and $\theta_{\mathrm{load}}$ is an overload threshold. This constraint filters unavailable or severely congested nodes, preventing routing failures or latency collapse under high system load.

\subsection{Semantic-Aware Routing via an SLM Task Router}
\label{subsec:slm_router}
Standard ant colony optimization (ACO) does not expose an explicit semantic interface, and thus tends to exhibit ``averaging'' behavior under mixed task streams, which amplifies cross-task interference in pheromone updates. AMRO-S introduces a lightweight small language model (SLM) as a semantic router that maps each query $q$ to a normalized task-mixture distribution over a predefined task set
\begin{equation}
\mathcal{T}=\{t_1,t_2,\ldots,t_k\}.
\end{equation}
Specifically, the router outputs a weight vector
\begin{equation}
\mathbf{w}(q)=\big(w_{t_1}(q),\ldots,w_{t_k}(q)\big), \qquad
w_t(q)\ge 0,\ \sum_{t\in\mathcal{T}} w_t(q)=1,
\label{eq:router_weight}
\end{equation}
where $k$ is the number of task types and $w_t(q)$ reflects the semantic attribution strength of $q$ to task $t$ (i.e., a task-mixture ratio). This vector serves as a \emph{semantic anchor} for query-conditioned fusion in subsequent routing components.

To obtain stable and controllable semantic signals, we construct an expert routing dataset
\begin{equation}
_{\mathrm{router}}=\{(q_i,\mathbf{w}_i^*)\}_{i=1}^{M},
\end{equation}
where $M$ is the number of training samples and $\mathbf{w}_i^*$ is the expert-annotated target distribution. We then perform supervised fine-tuning (SFT) by minimizing the KL divergence:
\begin{equation}
\mathcal{L}_{\mathrm{router}}=\sum_{i=1}^{M}\mathrm{KL}\!\left(\mathbf{w}_i^* \ \Vert\ \mathbf{w}(q_i)\right),
\label{eq:router_loss}
\end{equation}
where $\mathrm{KL}(\cdot\Vert\cdot)$ measures the discrepancy between two distributions. Since the router outputs only $\mathbf{w}(q)$ at inference time (instead of generating long-form reasoning), it provides a low-overhead semantic interface that enables routing decisions to explicitly adapt to mixed user intents.

\subsection{Multi-Task Pheromone Specialists and Query-Conditioned Fusion}
\label{subsec:pheromone_fusion}
To mitigate cross-task pheromone interference, AMRO-S does not maintain a single global pheromone matrix. Instead, for each task $t\in\mathcal{T}$, we maintain an independent pheromone specialist matrix $\tau^t$, where $\tau^t_{ij}$ accumulates the historical utility of choosing transition $(i\!\rightarrow\! j)$ under task $t$ (larger values indicate stronger preference). At inference time, we perform query-conditioned fusion via semantic superposition:
\begin{equation}
\tau^{(q)}_{ij}=\sum_{t\in\mathcal{T}} w_t(q)\cdot \tau^t_{ij},
\label{eq:pheromone_fusion}
\end{equation}
which yields a posterior pheromone $\tau^{(q)}$ aligned with the task mixture of query $q$. This factorize--fuse design (i) isolates task memories within $\{\tau^t\}$ to prevent contamination and (ii) enables smooth interpolation for mixed intents through the continuous weights $w_t(q)$.

Pheromone captures long-horizon experience but may respond slowly to instantaneous system dynamics such as congestion spikes. We therefore incorporate a task-aware heuristic term that combines capability priors with real-time signals. For node $j$ and task $t$, we define
\begin{equation}
\eta_j(t)=\lambda_A\cdot \widetilde{\mathrm{Ability}}[j][t]
+\lambda_L\cdot \widetilde{\Big(\frac{1}{\mathrm{Load}[j]+\epsilon}\Big)}
+\lambda_R\cdot \widetilde{\Big(\frac{1}{\mathrm{RT}[j]+\epsilon}\Big)},
\label{eq:heuristic_base}
\end{equation}
where $\mathrm{Ability}[j][t]$ is a task-specific capability prior estimated on a calibration set, $\mathrm{Load}[j]$ and $\mathrm{RT}[j]$ denote real-time load and response time, $\epsilon>0$ avoids division-by-zero, and $\lambda_A,\lambda_L,\lambda_R\ge 0$ control the relative contributions of the three signals. The operator $\widetilde{(\cdot)}$ denotes robust normalization (e.g., sliding-window min--max with quantile clipping) to align heterogeneous magnitudes. The query-conditioned heuristic is then
\begin{equation}
\eta^{(q)}_{j}=\sum_{t\in\mathcal{T}} w_t(q)\cdot \eta_j(t).
\label{eq:heuristic_fusion}
\end{equation}

Given $\tau^{(q)}$ and $\eta^{(q)}$, the transition probability from node $v_{l,i}$ to $v_{l+1,j}$ follows the standard ACO proportional rule:
\begin{equation}
p_{ij}(q)=
\frac{[\tau^{(q)}_{ij}]^{\alpha}\cdot[\eta^{(q)}_{j}]^{\beta}}
{\sum_{k\in \mathrm{Allowed}(l,i)}[\tau^{(q)}_{ik}]^{\alpha}\cdot[\eta^{(q)}_{k}]^{\beta}},
\label{eq:transition}
\end{equation}
where $\alpha,\beta>0$ control the importance of exploitation (pheromone) versus heuristic signals, and the denominator normalizes probabilities over feasible candidates $\mathrm{Allowed}(l,i)$. To prevent premature convergence caused by noisy early-stage updates, we adopt a minimum exploration safeguard: with probability $\gamma\in[0,1]$ we sample uniformly from $\mathrm{Allowed}(l,i)$; otherwise we sample according to $p_{ij}(q)$:
\begin{equation}
\Pr(\text{choose } j)=
\gamma\cdot \frac{1}{|\mathrm{Allowed}(l,i)|} + (1-\gamma)\cdot p_{ij}(q).
\label{eq:explore}
\end{equation}

\subsection{Offline Warm-up and Online Bypass Evolution}
\label{subsec:qgae}
AMRO-S adopts a two-stage optimization scheme---\emph{offline supervised warm-up} and \emph{online bypass evolution}---to achieve strong cold-start performance while retaining continual adaptation.

\paragraph{Offline supervised warm-up.}
For each task $t\in\mathcal{T}$, we optimize the corresponding specialist pheromone $\tau^t$ using labeled data. Given a sampled routing path $P$, we compute a task-dependent fitness score $f_t(P)$ from ground-truth signals (e.g., correctness or unit-test outcomes), where smaller $f_t(P)$ indicates a better path. We then apply the standard evaporation--reinforcement update:
\begin{equation}
\tau^t_{ij}\leftarrow (1-\rho)\cdot \tau^t_{ij}, \qquad
\tau^t_{ij}\leftarrow \tau^t_{ij}+\frac{Q}{f_t(P)+\epsilon},\ \ \forall (i,j)\in P,
\label{eq:offline_update}
\end{equation}
where $\rho\in(0,1)$ is the evaporation rate, $Q>0$ is a reinforcement scale, and $\epsilon>0$ stabilizes the update. Crucially, specialists for different tasks are trained independently, ensuring that $\{\tau^t\}$ converge to task-specific routing priors with reduced cross-task contamination.

\paragraph{Online bypass evolution with quality gating.}
In online deployment, most requests are unlabeled and the system must preserve low latency. We therefore decouple inference from learning. The serving (fast) path performs only: $\mathbf{w}(q)$ prediction $\rightarrow$ fusion of $\tau^{(q)}$ and $\eta^{(q)}$ $\rightarrow$ path sampling $\rightarrow$ agent execution and response, with \emph{no} on-the-fly updates. In parallel, we record a small fraction of requests at sampling rate $r$ into a FIFO buffer $\mathcal{B}$ as tuples $\langle q,P,\mathrm{output}\rangle$. When $|\mathcal{B}|=B$ (batch size), we trigger an asynchronous update. To control noise and prevent erroneous self-reinforcement, we introduce a lightweight LLM-Judge that outputs a binary gate:
\begin{equation}
g(q,P,\mathrm{output})\in\{0,1\},
\end{equation}
where $g=1$ indicates acceptable quality and $g=0$ discards the sample. For gated samples, we compute an online system fitness $f_{\mathrm{sys}}(P)$ from measurable overhead (e.g., weighted latency and token cost under the same accounting rule as in Eq.~\eqref{eq:cost_decompose}) and update specialists proportionally to the router weights:
\begin{equation}
\tau^t_{ij}\leftarrow (1-\rho)\cdot \tau^t_{ij}
+ w_t(q)\cdot \frac{Q}{f_{\mathrm{sys}}(P)+\epsilon},
\qquad \forall (i,j)\in P,\ \forall t\in\mathcal{T}.
\label{eq:online_update}
\end{equation}
This design ensures that (i) pheromone is reinforced only by high-quality trajectories (via gating), and (ii) update strength aligns with the semantic mixture $w_t(q)$, enabling continual, controlled, and task-decoupled routing adaptation without introducing additional serving overhead.

\begin{table*}[h]
\centering
\caption{\textbf{Performance comparison on five benchmarks.} Mul. and Rout. denote Multi-Agent and Dynamic Routing, respectively. The LLM Pool* comprises cost-effective models (GPT-4o-mini, Gemini-1.5-flash, Claude-3.5-haiku, Llama-3.1-70b).}
\label{tab:llm_performance}
\resizebox{0.9\textwidth}{!}{
\begin{tabular}{llcccccccc}
\toprule
\textbf{Method} & \textbf{LLM} & \textbf{Mul.} & \textbf{Rout.} & \textbf{MMLU} & \textbf{GSM8K} & \textbf{MATH} & \textbf{HumanEval} & \textbf{MBPP} & \textbf{Avg.} \\ 
\midrule
\multicolumn{10}{c}{\textit{\textbf{Reference Models}}} \\
Vanilla (Ref) & GPT-4o & N & N & 88.7 & 96.1 & 76.6 & 90.2 & 87.2 & 87.76 \\
Vanilla (Ref) & Claude-3.5-Sonnet & N & N & 88.3 & 96.4 & 78.0 & 93.7 & 89.1 & 89.10 \\
\midrule
\multicolumn{10}{c}{\textit{\textbf{Single-Agent Baselines}}} \\
Vanilla & GPT-4o-mini & N & N & 77.81 & 93.17 & 66.09 & 85.71 & 72.2 & 79.00 \\
Vanilla & Gemini-1.5-flash & N & N & 80.04 & 92.67 & 74.39 & 82.61 & 73.0 & 80.54 \\
Vanilla & Claude-3.5-haiku & N & N & 78.5 & 91.8 & 68.2 & 86.4 & 74.5 & 79.88 \\
Vanilla & Llama-3.1-70b & N & N & 82.3 & 94.1 & 68.0 & 80.5 & 71.8 & 79.34 \\
\midrule
\multicolumn{10}{c}{\textit{\textbf{Chain-based Reasoning}}} \\
CoT & GPT-4o-mini & N & N & 78.43 & 93.68 & 67.24 & 86.69 & 69.6 & 79.13 \\
ToT (Tree) & GPT-4o-mini & N & N & 79.1 & 94.2 & 71.5 & 85.2 & 72.8 & 80.56 \\
GoT (Graph) & GPT-4o-mini & N & N & 79.5 & 94.5 & 72.1 & 85.8 & 73.2 & 81.02 \\
AoT (Atom) & GPT-4o-mini & N & N & 80.2 & 95.1 & 72.8 & 87.5 & 74.0 & 81.92 \\
\midrule
\multicolumn{10}{c}{\textit{\textbf{Multi-Agent Baselines}}} \\
LLM-Debate & GPT-4o-mini & Y & N & 81.04 & 94.66 & 64.68 & 84.38 & 73.6 & 79.67 \\
GPTSwarm & GPT-4o-mini & Y & N & 82.8 & 94.66 & 68.85 & 86.28 & 75.4 & 81.60 \\
AFlow & GPT-4o-mini & Y & N & 83.1 & 92.3 & 73.35 & 90.06 & 82.2 & 84.20 \\
AFlow & Gemini-1.5-flash & Y & N & 82.35 & 94.91 & 72.7 & 85.69 & 76.0 & 82.33 \\
\midrule
\multicolumn{10}{c}{\textit{\textbf{Routing Methods}}} \\
RouteLLM & LLM Pool* & N & Y & 81.04 & 93.42 & 71.29 & 83.85 & 72.6 & 80.44 \\
RouterDC & LLM Pool* & N & Y & 82.01 & 93.68 & 73.46 & 87.75 & 75.2 & 82.42 \\
MasRouter & LLM Pool* & Y & Y & 84.25 & 95.45 & 75.42 & 90.62 & 84.0 & 85.93 \\
\midrule
\rowcolor{gray!15}
\textbf{AMRO-S (Ours)} & \textbf{LLM Pool*} & \textbf{Y} & \textbf{Y} & \textbf{86.1} & \textbf{96.4} & \textbf{78.15} & \textbf{92.2} & \textbf{86.3} & \textbf{87.83} \\
\bottomrule
\end{tabular}
}
\end{table*}

\section{Experiments}
\label{Experiments}
In this section, we conduct extensive experiments on five public benchmarks to systematically evaluate the effectiveness, efficiency, and interpretability of our proposed framework, AMRO-S. Specifically, we aim to address the following research questions:

\begin{itemize}
    \item \textbf{RQ1:} Does AMRO-S outperform state-of-the-art single-agent baselines and existing routing methods across diverse reasoning and coding tasks?
    \item \textbf{RQ2:} Can AMRO-S be seamlessly integrated into existing multi-agent frameworks to improve the quality--cost trade-off?
    \item \textbf{RQ3:} Are the key components of AMRO-S, particularly the SFT-enhanced SLM router and the pheromone-based routing mechanism, effective and necessary?
    \item \textbf{RQ4:} How does AMRO-S perform under high-concurrency scenarios? Can it maintain stability and low latency under extreme system loads?
    \item \textbf{RQ5:} Does the evolution of pheromone specialists provide traceable and semantically meaningful evidence for routing decisions?
\end{itemize}

\subsection{Experimental Setup}
\label{Experimental Setup}

\textbf{Models.}
To construct a heterogeneous and cost-effective agent pool, we selected four representative models spanning proprietary and open-source families: gpt-4o-mini, gemini-1.5-flash, claude-3.5-haiku, and llama-3.1-70b. This selection ensures diversity in reasoning patterns and pricing structures while maintaining high accessibility. For the semantic router backbone, we employed lightweight Small Language Models, specifically Llama-3.2-1B-Instruct and Qwen2.5-1.5B, to minimize routing overhead while maintaining adequate intent recognition capabilities. Additionally, state-of-the-art models including GPT-4o and Claude-3.5-Sonnet serve as high-capability single-agent baselines for performance benchmarking.

\textbf{Dataset and Benchmarks.}
We validated the model on five public datasets, including GSM8K \cite{cobbe2021training}, MMLU \cite{hendrycks2020measuring}, MATH \cite{hendrycks2024measuring}, HumanEval \cite{chen2021evaluating}, and MBPP \cite{austin2021program}. GSM8K is a dataset of 8.5K high-quality, linguistically diverse primary school math word problems, while MMLU covers 57 distinct cate- gories ranging from basic knowledge to advanced professional disciplines. MATH, a math competition problem dataset, provides complete step-by-step solutions for each problem to train models to generate answer derivation processes and explanations. HumanEval is designed for evaluating code- generation models, containing programming problems with function signatures, docstrings, function bodies, and multiple unit tests, whereas MBPP consists of short Python programs crowdsourced from individuals with basic Python knowledge. These datasets collectively enable comprehensive assessment of the model’s performance across mathematical reasoning, domain-specific knowledge, code generation, and problem-solving capabilities.

\textbf{Baselines.}
To ensure a fair and reproducible comparison, we report results only for the methods included in our tables. The baselines cover four types of approaches. 
Vanilla refers to direct single-model inference without structured reasoning or multi-agent collaboration. 
Chain-based reasoning methods include Chain-of-Thought (CoT) \cite{wei2022chain}, Tree-of-Thoughts (ToT) \cite{yao2023tree}, Graph-of-Thoughts (GoT) \cite{besta2024graph}, and Algorithm-of-Thoughts (AoT) \cite{teng2025atom}. 
Multi-agent baselines perform collaborative problem solving without explicit routing mechanisms, including LLM-Debate, GPTSwarm \cite{zhuge2024gptswarm}, and AFlow \cite{zhang2024aflow}. 
Routing methods introduce dynamic selection over candidate models or paths, represented by RouteLLM \cite{ong2024routellm}, RouterDC \cite{chen2024routerdc}, and the multi-agent router MasRouter \cite{yue2025masrouter}. 
In addition, to evaluate plug-and-play adaptability, we integrate AMRO-S into three representative MAS frameworks, MacNet \cite{qian2024scaling}, GPTSwarm, and HEnRY \cite{lacavalla2024henry}, and compare against their original routing policies while keeping the execution workflow unchanged.

\textbf{Implementation Details.}
We adopt Pass@1 as the primary evaluation metric, where mathematical reasoning is assessed via Exact Match (EM) and coding performance is validated through unit test execution requiring a 100\% pass rate. To isolate the effects of routing logic in our AMRO-S framework, we enforce a strict unified inference budget by standardizing the maximum interaction turns ($T_{max}$) and total agent invocations ($I_{max}$) per query, ensuring performance gains stem from superior routing rather than extended generation steps. Cost analysis is performed using a high-fidelity attribution model based on official API pricing, aggregating token consumption across router overhead, intermediate reasoning, and final execution. For Semantic Router Construction, we curated a specialized training set $\mathcal{D}_{router}$ of 3,000 instructions via GPT-4o using an adaptive topic-selection strategy to ensure intent diversity. We fine-tuned Llama-3.2-1B-Instruct and Qwen2.5-1.5B by minimizing the KL divergence between predicted and target distributions to achieve effective knowledge distillation. Training was conducted for 3 epochs with a batch size of 64 and a learning rate of $2 \times 10^{-5}$ (cosine decay, 0.03 warmup) using the AdamW optimizer (weight decay 0.01). All experiments were executed on a single NVIDIA A100 (80GB) GPU, with router training typically concluding within 30 minutes.

\subsection{Main Results}
\label{Main Results}

\begin{table*}[t] 
\centering
\caption{\textbf{Adaptability and Cost-Efficiency Evaluation on Existing Multi-Agent Frameworks.} We integrate the classifier-based \textit{MasRouter} and our pheromone-based \textit{AMRO-S} into three representative frameworks (MacNet, GPTSwarm, HEnRY). GPT: GPT-4o-mini; Gemini: Gemini-1.5-flash.}
\label{tab:adaptability_experiment}
\resizebox{0.8\textwidth}{!}{
\begin{tabular}{lll ccc ccc ccc}
\toprule
\multirow{2}{*}{\textbf{Framework}} & \multirow{2}{*}{\textbf{Dataset}} & \multirow{2}{*}{\textbf{Model}} & \multicolumn{2}{c}{\textbf{Original}} & & \multicolumn{2}{c}{\textbf{+MasRouter}} & & \multicolumn{2}{c}{\textbf{+AMRO-S (Ours)}} \\
\cmidrule{4-5} \cmidrule{7-8} \cmidrule{10-11}
 & & & \textbf{Acc.} & \textbf{Cost} & & \textbf{Acc.} & \textbf{Cost} & & \textbf{Acc.} & \textbf{Cost} \\
\midrule
\multirow{6}{*}{MacNet} 
 & \multirow{2}{*}{MMLU} & GPT & 82.98\% & 7.81 & & 83.25\% & 7.65 & & \textbf{83.50\%} & \textbf{7.50} \\
 & & Gemini & 81.74\% & 8.90 & & 81.95\% & 8.65 & & \textbf{82.10\%} & \textbf{8.40} \\
\cmidrule{2-11}
 & \multirow{2}{*}{HumanEval} & GPT & 86.82\% & 0.49 & & 87.20\% & 0.48 & & \textbf{87.50\%} & \textbf{0.47} \\
 & & Gemini & 88.72\% & 0.53 & & 88.85\% & 0.52 & & \textbf{89.00\%} & \textbf{0.50} \\
\cmidrule{2-11}
 & \multirow{2}{*}{GSM8K} & GPT & 94.69\% & 2.14 & & 94.85\% & 2.08 & & \textbf{95.00\%} & \textbf{2.00} \\
 & & Gemini & 94.31\% & 2.20 & & 94.42\% & 2.12 & & \textbf{94.50\%} & \textbf{2.05} \\
\midrule
\multirow{6}{*}{GPTSwarm} 
 & \multirow{2}{*}{MMLU} & GPT & 83.00\% & 8.00 & & 83.80\% & 7.75 & & \textbf{84.20\%} & \textbf{7.40} \\
 & & Gemini & 81.50\% & 8.90 & & 82.40\% & 8.60 & & \textbf{82.90\%} & \textbf{8.30} \\
\cmidrule{2-11}
 & \multirow{2}{*}{HumanEval} & GPT & 87.30\% & 0.51 & & 88.20\% & 0.49 & & \textbf{88.80\%} & \textbf{0.47} \\
 & & Gemini & 88.50\% & 0.55 & & 88.80\% & 0.53 & & \textbf{89.10\%} & \textbf{0.50} \\
\cmidrule{2-11}
 & \multirow{2}{*}{GSM8K} & GPT & 94.80\% & 2.10 & & 94.92\% & 2.00 & & \textbf{95.00\%} & \textbf{1.90} \\
 & & Gemini & 94.30\% & 2.15 & & 94.55\% & 2.10 & & \textbf{94.70\%} & \textbf{2.05} \\
\midrule
\multirow{6}{*}{HEnRY} 
 & \multirow{2}{*}{MMLU} & GPT & 82.80\% & 8.30 & & 83.40\% & 8.00 & & \textbf{83.80\%} & \textbf{7.70} \\
 & & Gemini & 81.20\% & 9.00 & & 82.10\% & 8.75 & & \textbf{82.70\%} & \textbf{8.50} \\
\cmidrule{2-11}
 & \multirow{2}{*}{HumanEval} & GPT & 87.10\% & 0.52 & & 87.50\% & 0.49 & & \textbf{87.80\%} & \textbf{0.46} \\
 & & Gemini & 88.00\% & 0.55 & & 88.60\% & 0.52 & & \textbf{88.50\%} & \textbf{0.49} \\
\cmidrule{2-11}
 & \multirow{2}{*}{GSM8K} & GPT & 94.50\% & 2.15 & & 94.65\% & 2.05 & & \textbf{94.80\%} & \textbf{1.90} \\
 & & Gemini & 94.00\% & 2.20 & & 94.30\% & 2.12 & & \textbf{94.50\%} & \textbf{2.05} \\
\bottomrule
\end{tabular}
}
\end{table*}

\textbf{Answer to RQ1:}
To examine whether AMRO-S outperforms state-of-the-art baselines, we construct a unified evaluation protocol across five public benchmarks and compare single-model baselines, chain-based reasoning strategies, multi-agent systems without routing, and representative routing methods under the same LLM Pool and consistent inference-budget constraints. Table~\ref{tab:llm_performance} reports the pass@1 performance on MMLU, GSM8K, MATH, HumanEval, and MBPP. Under this unified setup, AMRO-S achieves the best overall performance with an average score of 87.83. Compared with MasRouter, the strongest multi-agent routing baseline, AMRO-S raises the average score from 85.93 to 87.83. The improvements are more pronounced on harder reasoning and coding tasks: the score on MATH increases from 75.42 to 78.15, and the score on MBPP increases from 84.0 to 86.3. These results suggest that AMRO-S more reliably aligns task semantics, path structure, and model capability under mixed workloads, preventing the gains of collaboration from being offset by capability mismatch and redundant invocations. In contrast to static multi-agent topologies that may become inefficient and unstable when task distributions shift, AMRO-S combines SLM-based semantic routing, task-isolated pheromone specialists, and quality-gated online evolution to translate collaboration into consistent cross-task improvements.

\textbf{Answer to RQ2:}
Following the unified inference-budget constraints and cost accounting described in the experimental setup, we plug AMRO-S into MacNet, GPTSwarm, and HEnRY while keeping their agent composition and execution workflow unchanged, and only replacing the path-selection policy. Table~\ref{tab:adaptability_experiment} reports the resulting accuracy and inference cost under both gpt-4o-mini and gemini-1.5-flash backbones. Across all three frameworks and both backbone configurations, the +AMRO setting consistently yields higher accuracy than the original frameworks and the +MasRouter variants. For instance, on MacNet with gpt-4o-mini, AMRO-S improves MMLU accuracy from 82.98\% to 83.50\%. Meanwhile, AMRO-S achieves these gains with the lowest inference cost, reflecting a better cost--quality trade-off. On GSM8K under MacNet, the cost decreases from \$2.14 in the original setting to \$2.00 with AMRO-S, indicating that the routing mechanism learns more economical path preferences from feedback signals. Overall, these results support that AMRO-S functions as a portable and independent semantic-driven routing layer that can enhance diverse MAS architectures without imposing additional resource burdens.

\begin{table*}[h]
\centering
\caption{\textbf{Ablation Study on Router Components.} Analysis of single-agent baselines versus different router backbones and strategies.}
\label{tab:ablation_new}
\resizebox{\textwidth}{!}{
\begin{tabular}{lllccccccccc}
\toprule
\textbf{ID} & \textbf{Setting} & \textbf{Router Backbone} & \textbf{Strategy} & \textbf{Mul.} & \textbf{Rout.} & \textbf{GSM8K} & \textbf{MATH} & \textbf{MMLU} & \textbf{HumanEval} & \textbf{MBPP} & \textbf{Avg.} \\ 
\midrule
\multicolumn{12}{c}{\textit{\textbf{Constituents (Single-Agent Baselines)}}} \\
(A) & Single & Gpt-4o-mini & N/A & $\times$ & $\times$ & 93.17 & 66.09 & 77.81 & 85.71 & 72.20 & 79.00 \\
(B) & Single & Gemini-1.5-flash & N/A & $\times$ & $\times$ & 92.67 & 74.39 & 80.04 & 82.61 & 73.00 & 80.54 \\
(C) & Single & Claude-3.5-haiku & N/A & $\times$ & $\times$ & 91.80 & 68.20 & 78.50 & 86.40 & 74.50 & 79.88 \\
(D) & Single & Llama-3.1-70b & N/A & $\times$ & $\times$ & 94.10 & 68.00 & 82.30 & 80.50 & 71.80 & 79.34 \\
\midrule
\multicolumn{12}{c}{\textit{\textbf{Router and Strategy Ablation}}} \\
(E) & w/o Routing & Random & Random & $\checkmark$ & $\times$ & 92.90 & 69.10 & 79.60 & 83.80 & 72.80 & 79.64 \\
(F) & w/o SFT & Llama-3.2-1B & w/o SFT & $\checkmark$ & $\checkmark$ & 94.50 & 73.20 & 82.50 & 88.40 & 78.50 & 83.42 \\
(G) & w/o SFT & GPT-4o-mini & w/o SFT & $\checkmark$ & $\checkmark$ & 95.80 & 76.50 & 85.20 & 90.80 & 84.10 & 86.48 \\
(H) & w/ SFT & Qwen2.5-1.5B & w/ SFT & $\checkmark$ & $\checkmark$ & 96.20 & 77.90 & 85.95 & 92.00 & 86.10 & 87.63 \\
\midrule
\rowcolor{gray!15}
(I) & \textbf{AMRO-S} & \textbf{Llama-3.2-1B} & \textbf{w/ SFT} & $\checkmark$ & $\checkmark$ & \textbf{96.40} & \textbf{78.15} & \textbf{86.10} & \textbf{92.20} & \textbf{86.30} & \textbf{87.83} \\
\bottomrule
\end{tabular}
}
\end{table*}

\subsection{Ablation Study}
\label{Ablation Study}
\begin{table}[h]
\centering
\caption{\textbf{SLM Intent Recognition Accuracy.} Comparison of zero-shot baselines and fine-tuned models (SFT) across different intents.}
\label{tab:slm_results}
\resizebox{0.5\textwidth}{!}{
\begin{tabular}{lccccc}
\toprule
\textbf{Model} & \textbf{SFT} & \textbf{Math} & \textbf{Code} & \textbf{General} & \textbf{Avg.} \\ \midrule
Llama-3.2-1B-Instruct & $\times$ & 78.50\% & 82.10\% & 85.40\% & 82.00\% \\
Qwen2.5-1.5B          & $\times$ & 84.20\% & 88.50\% & 89.10\% & 87.26\% \\
GPT-4o-mini           & $\times$ & 95.20\% & 96.10\% & 96.20\% & 95.83\% \\ \midrule
Qwen2.5-1.5B          & $\checkmark$ & 97.90\% & 98.20\% & 97.50\% & 97.86\% \\
\rowcolor{gray!15}
Llama-3.2-1B-Instruct & $\checkmark$ & \textbf{98.10\%} & \textbf{97.90\%} & \textbf{97.80\%} & \textbf{97.93\%} \\ \bottomrule
\end{tabular}
}
\end{table}
\textbf{Answer to RQ3:}
Table~\ref{tab:ablation_new} reports an end-to-end ablation over routing configurations, covering single-agent constituents, multi-agent execution without routing, and router variants that differ in backbone capacity and whether supervised fine-tuning is applied. The results indicate that multi-agent collaboration alone is not sufficient to yield stable gains. When execution paths are selected randomly, the overall average remains at 79.64, which is close to several single-agent baselines, suggesting that capability mismatch and redundant calls can dilute the benefits of collaboration. Enabling routing consistently improves performance, and the routing quality is shaped by both router capacity and supervision. A compact router without SFT already provides a noticeable improvement over random routing, reaching an average of 83.42 with Llama-3.2-1B, but still leaves a clear gap to the full system. Increasing the router capacity without SFT further raises the average to 86.48, showing that a stronger router can partially compensate for missing alignment. Applying SFT yields larger and more stable gains even with compact routers, lifting the average to 87.63 with Qwen2.5-1.5B and achieving the best performance of 87.83 in AMRO-S with the SFT-enhanced Llama-3.2-1B router. Overall, these results support that semantic routing alignment and pheromone-guided path optimization jointly contribute to consistent multi-task improvements.

Table~\ref{tab:slm_results} evaluates the SLM router independently on intent recognition across math, code, and general queries, providing a direct measure of the semantic signal quality that drives downstream routing. Without SFT, lightweight routers show noticeable intent classification gaps, which can propagate into unstable routing decisions under mixed workloads, whereas relying on a stronger LLM router yields higher accuracy but is less desirable for cost-sensitive deployment. After SFT, both Qwen2.5-1.5B and Llama-3.2-1B achieve near-saturated intent recognition, reaching 97.86\% and 97.93\% overall accuracy, with consistently high scores across all intent categories. This indicates that SFT effectively anchors compact routers into reliable task-intent predictors, offering a low-cost yet high-precision semantic interface for pheromone fusion and subsequent path selection.

\subsection{Efficiency and Scalability Analysis}
\label{Efficiency and Scalability Analysis}

\textbf{Answer to RQ4:}
To evaluate whether AMRO-S remains efficient and stable under high concurrency, we conduct a stress test by scaling the concurrency level from 20 to 1000 processes and comparing against a weighted round-robin baseline. In this experiment, processes denotes the number of concurrent workers that execute requests in parallel, and all settings process the same fixed set of queries with identical prompts and termination criteria, so that the workload scale and difficulty remain unchanged across concurrency levels. Table~\ref{tab:stress_test} reports the end-to-end wall-clock time for completing the full workload, where speedup is computed relative to the 20-process setting under the same workload. We also report pass@1 accuracy measured on GSM8K for each concurrency level, using the same evaluation script and aggregation protocol for AMRO-S and WRR.

As concurrency increases, AMRO-S exhibits clear scalability: the total runtime decreases from 3849.60 seconds at 20 processes to 823.21 seconds at 1000 processes, corresponding to a 4.7$\times$ speedup. Importantly, the accuracy remains stable, staying within 96.10\% to 96.40\% across all concurrency levels, indicating that AMRO-S preserves capability--task matching under heavy load. In contrast, WRR shows progressively degraded accuracy as concurrency increases, dropping from 96.00\% at 20 processes to 88.20\% at 1000 processes, suggesting that naive load-balancing can break semantic-aware routing behavior under extreme system pressure. Overall, these results demonstrate that AMRO-S achieves a favorable throughput--quality trade-off in highly parallel settings, while maintaining stable routing decisions under dynamic, high-concurrency workloads.

\begin{table}[htbp]
\centering
\caption{\textbf{Stress test under varying concurrency.} ``Processes'' denotes the number of parallel workers; speedup is relative to 20 processes.}
\label{tab:stress_test}
\resizebox{\linewidth}{!}{
\begin{tabular}{lcccccc}
\toprule
\textbf{Processes} & \textbf{20} & \textbf{50} & \textbf{100} & \textbf{200} & \textbf{500} & \textbf{1000} \\ 
\midrule
\multicolumn{7}{c}{\textit{\textbf{AMRO-S Efficiency Metrics}}} \\ 
Time (s) & 3849.60 & 2430.40 & 1863.75 & 1382.90 & 1062.30 & 823.21 \\
Time (min) & 64.16 & 40.51 & 31.06 & 23.05 & 17.71 & 13.72 \\
Speedup & 1.0$\times$ & 1.6$\times$ & 2.1$\times$ & 2.8$\times$ & 3.6$\times$ & \textbf{4.7}$\times$ \\
\midrule
\multicolumn{7}{c}{\textit{\textbf{Accuracy Comparison}}} \\ 
WRR(Baseline) & 96.00\% & 95.80\% & 95.20\% & 93.50\% & 91.50\% & 88.20\% \\
\rowcolor{gray!15}
\textbf{AMRO-S (Ours)} & \textbf{96.10\%} & \textbf{96.20\%} & \textbf{96.25\%} & \textbf{96.30\%} & \textbf{96.40\%} & \textbf{96.40\%} \\
\bottomrule
\end{tabular}
}
\end{table}

\subsection{Interpretability Analysis: Visualizing the Pheromone Specialists}
\label{Interpretability Analysis}

\begin{figure*}[h]
    \centering
    \includegraphics[width=0.95\linewidth]{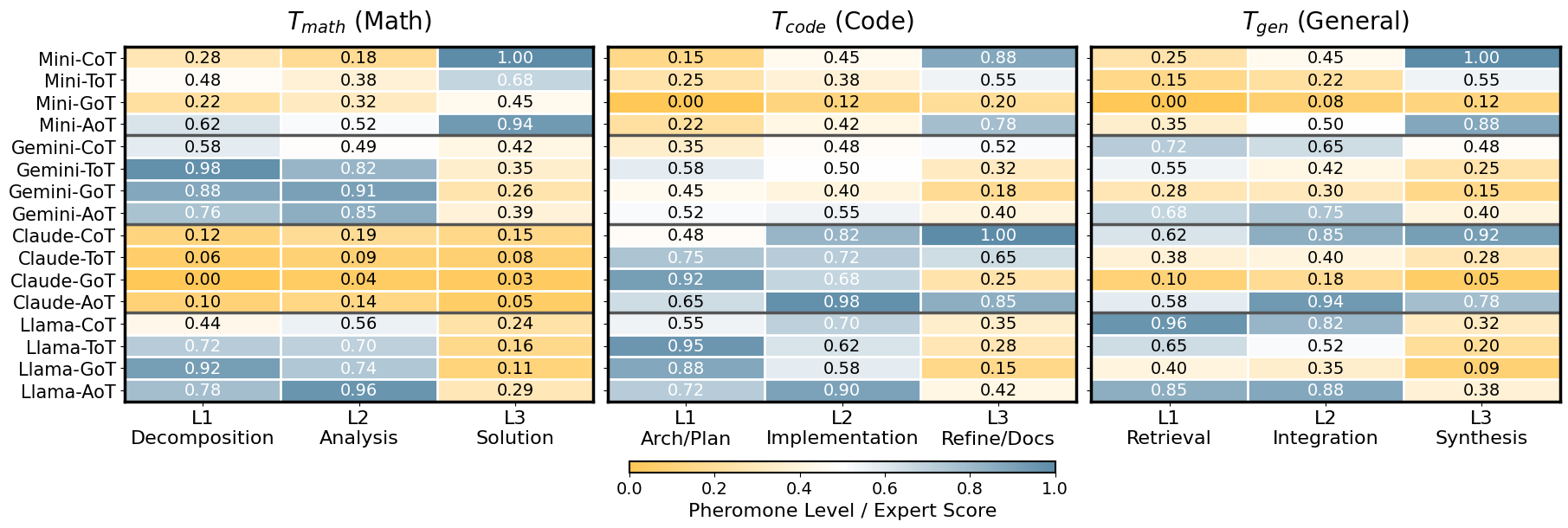}
    \caption{Converged pheromone specialists of AMRO-S for three domains: mathematical reasoning $T_{math}$, code generation $T_{code}$, and general reasoning $T_{gen}$. Color intensity indicates the learned routing preference, where deeper teal denotes stronger preference and lighter tones denote weaker preference.}
    \label{fig:heatmap}
\end{figure*}

\noindent\textbf{Answer to RQ5:}
To examine the decision logic of AMRO-S and assess its interpretability, we visualize the converged task-specific pheromone specialists after training. Figure~\ref{fig:heatmap} presents three specialists corresponding to mathematical reasoning $T_{math}$, code generation $T_{code}$, and general reasoning $T_{gen}$, where each heatmap encodes the learned preference over stage-to-stage transitions. Rather than acting as an opaque router, these pheromone distributions provide explicit and traceable evidence of how the quality-gated asynchronous evolution mechanism automatically discovers adaptive, task-specific collaboration topologies. Specifically, for $T_{code}$, pheromone intensity concentrates on a small subset of transitions in the later stages, which occurs because code generation is sensitive to syntax and logical edge cases, making the final implementation stage a critical bottleneck. The binary quality gate effectively filters out trajectories that fail unit tests, guiding the system to stably converge on reliable coding backbones in the final layer to ensure executability. Conversely, for $T_{math}$, the preference distribution exhibits a clear temporal variance as earlier stages emphasize candidates that support problem decomposition while later stages shift toward candidates yielding precise final calculations. This implicitly learned division of labor arises from the strict sequential dependency of mathematical reasoning, as trajectories lacking early strategic planning or late-stage precision fail to pass the exact-match evaluation. For $T_{gen}$, the system optimizes over the joint space of reasoning strategy and execution role, resulting in a more distributed routing pattern that balances answer quality against token overheads. Collectively, these task-specific pheromone patterns demonstrate that AMRO-S functions as an automated workflow discoverer that can autonomously identify and optimize suitable routing trajectories for heterogeneous workloads.

\section{Conclusion}
In this paper, we propose AMRO-S, an efficient and interpretable routing framework for heterogeneous LLM-based MAS under mixed intents and dynamic serving constraints. AMRO-S models MAS routing as a semantic-aware path search on a layered directed graph and combines an SFT-enhanced small language model for intent inference, task-specific pheromone specialists for task-isolated routing memory, and quality-gated asynchronous updates for controlled online refinement. Experiments on five benchmarks and integration evaluations on existing MAS frameworks show that AMRO-S consistently improves the quality--cost trade-off over strong routing baselines. High-concurrency stress tests further demonstrate favorable scalability and stable accuracy, while pheromone specialists provide traceable evidence for path selection, supporting transparent and deployable agent orchestration.

\section*{Acknowledgments}
This work was supported by the National Natural Science Foundation of China (NSFC) under the General Program (Grant No. 62572104).

This article used large language models (such as ChatGPT) as an auxiliary tool in the language polishing process, but did not use them in research conception and academic content generation.

\bibliographystyle{IEEEtran}
\bibliography{main}

@article{wang2024survey,
  title={A survey on large language model based autonomous agents},
  author={Wang, Lei and Ma, Chen and Feng, Xueyang and Zhang, Zeyu and Yang, Hao and Zhang, Jingsen and Chen, Zhiyuan and Tang, Jiakai and Chen, Xu and Lin, Yankai and others},
  journal={Frontiers of Computer Science},
  volume={18},
  number={6},
  pages={186345},
  year={2024},
  publisher={Springer}
}

@article{wei2022chain,
  title={Chain-of-thought prompting elicits reasoning in large language models},
  author={Wei, Jason and Wang, Xuezhi and Schuurmans, Dale and Bosma, Maarten and Xia, Fei and Chi, Ed and Le, Quoc V and Zhou, Denny and others},
  journal={Advances in neural information processing systems},
  volume={35},
  pages={24824--24837},
  year={2022}
}

@article{chen2024routerdc,
  title={Routerdc: Query-based router by dual contrastive learning for assembling large language models},
  author={Chen, Shuhao and Jiang, Weisen and Lin, Baijiong and Kwok, James and Zhang, Yu},
  journal={Advances in Neural Information Processing Systems},
  volume={37},
  pages={66305--66328},
  year={2024}
}

@article{dorri2018multi,
  title={Multi-agent systems: A survey},
  author={Dorri, Ali and Kanhere, Salil S and Jurdak, Raja},
  journal={Ieee Access},
  volume={6},
  pages={28573--28593},
  year={2018},
}

@article{balaji2010introduction,
  title={An introduction to multi-agent systems},
  author={Balaji, Parasumanna Gokulan and Srinivasan, Dipti},
  journal={Innovations in multi-agent systems and applications-1},
  pages={1--27},
  year={2010},
}

@article{kasneci2023chatgpt,
  title={ChatGPT for good? On opportunities and challenges of large language models for education},
  author={Kasneci, Enkelejda and Se{\ss}ler, Kathrin and K{\"u}chemann, Stefan and Bannert, Maria and Dementieva, Daryna and Fischer, Frank and Gasser, Urs and Groh, Georg and G{\"u}nnemann, Stephan and H{\"u}llermeier, Eyke and others},
  journal={Learning and individual differences},
  volume={103},
  pages={102274},
  year={2023},

}

@article{li2024survey,
  title={A survey on LLM-based multi-agent systems: workflow, infrastructure, and challenges},
  author={Li, Xinyi and Wang, Sai and Zeng, Siqi and Wu, Yu and Yang, Yi},
  journal={Vicinagearth},
  volume={1},
  number={1},
  pages={9},
  year={2024},
 
}

@article{han2024llm,
  title={LLM multi-agent systems: Challenges and open problems},
  author={Han, Shanshan and Zhang, Qifan and Yao, Yuhang and Jin, Weizhao and Xu, Zhaozhuo and He, Chaoyang},
  journal={arXiv preprint arXiv:2402.03578},
  year={2024}
}

@article{lu2023routing,
  title={Routing to the expert: Efficient reward-guided ensemble of large language models},
  author={Lu, Keming and Yuan, Hongyi and Lin, Runji and Lin, Junyang and Yuan, Zheng and Zhou, Chang and Zhou, Jingren},
  journal={arXiv preprint arXiv:2311.08692},
  year={2023}
}

@article{liang2023encouraging,
  title={Encouraging Divergent Thinking in Large Language Models through Multi-Agent Debate},
  author={Liang, Tian and He, Zhiwei and Jiao, Wenxiang and Wang, Xing and Wang, Yan and Wang, Rui and Yang, Yujiu and Tu, Zhaopeng and Shi, Shuming},
  journal={arXiv preprint arXiv:2305.19118},
  year={2023}
}

@article{ong2024routellm,
  title={Routellm: Learning to route llms with preference data},
  author={Ong, Isaac and Almahairi, Amjad and Wu, Vincent and Chiang, Wei-Lin and Wu, Tianhao and Gonzalez, Joseph E and Kadous, M Waleed and Stoica, Ion},
  journal={arXiv preprint arXiv:2406.18665},
  year={2024}
}

@article{yue2025masrouter,
  title={Masrouter: Learning to route llms for multi-agent systems},
  author={Yue, Yanwei and Zhang, Guibin and Liu, Boyang and Wan, Guancheng and Wang, Kun and Cheng, Dawei and Qi, Yiyan},
  journal={arXiv preprint arXiv:2502.11133},
  year={2025}
}

@article{yahia2023path,
  title={Path planning optimization in unmanned aerial vehicles using meta-heuristic algorithms: A systematic review},
  author={Yahia, Hazha Saeed and Mohammed, Amin Salih},
  journal={Environmental Monitoring and Assessment},
  volume={195},
  number={1},
  pages={30},
  year={2023},
 
}

@article{sivanandam2008genetic,
  title={Genetic algorithm optimization problems},
  author={Sivanandam, SN and Deepa, SN and Sivanandam, SN and Deepa, SN},
  journal={Introduction to genetic algorithms},
  pages={165--209},
  year={2008},
}

@article{rutenbar1989simulated,
  title={Simulated annealing algorithms: An overview},
  author={Rutenbar, Rob A},
  journal={IEEE Circuits and Devices magazine},
  volume={5},
  number={1},
  pages={19--26},
  year={1989},

}

@article{wang2018particle,
  title={Particle swarm optimization algorithm: an overview},
  author={Wang, Dongshu and Tan, Dapei and Liu, Lei},
  journal={Soft computing},
  volume={22},
  number={2},
  pages={387--408},
  year={2018},
 
}

@article{gad2022particle1,
  title={Particle swarm optimization algorithm and its applications: a systematic review},
  author={Gad, Ahmed G},
  journal={Archives of computational methods in engineering},
  volume={29},
  number={5},
  pages={2531--2561},
  year={2022},

}

@article{cui2024multi,
  title={Multi-strategy adaptable ant colony optimization algorithm and its application in robot path planning},
  author={Cui, Junguo and Wu, Lei and Huang, Xiaodong and Xu, Dengpan and Liu, Chao and Xiao, Wensheng},
  journal={Knowledge-Based Systems},
  volume={288},
  pages={111459},
  year={2024},
}

@article{blum2005ant,
  title={Ant colony optimization: Introduction and recent trends},
  author={Blum, Christian},
  journal={Physics of Life reviews},
  volume={2},
  number={4},
  pages={353--373},
  year={2005},
}

@article{scianna2024addaco,
  title={The AddACO: A bio-inspired modified version of the ant colony optimization algorithm to solve travel salesman problems},
  author={Scianna, Marco},
  journal={Mathematics and computers in simulation},
  volume={218},
  pages={357--382},
  year={2024},
  
}

@article{liang2024enhanced,
  title={An enhanced ant colony optimization algorithm for global path planning of deep-sea mining vehicles},
  author={Liang, Weixing and Lou, Min and Chen, Zhangxing and Qin, Huiyang and Zhang, Chen and Cui, Chengwei and Wang, Yangyang},
  journal={Ocean Engineering},
  volume={301},
  pages={117415},
  year={2024},
}

@article{si2024novel,
  title={A novel parallel ant colony optimization algorithm for mobile robot path planning.},
  author={Si, Jian and Bao, Xiaoguang},

  volume={21},
  number={2},
  pages={2568--2586},
  year={2024}
}

@article{tan2021comprehensive,
  title={A comprehensive review of coverage path planning in robotics using classical and heuristic algorithms},
  author={Tan, Chee Sheng and Mohd-Mokhtar, Rosmiwati and Arshad, Mohd Rizal},
  journal={IEEE Access},
  volume={9},
  pages={119310--119342},
  year={2021},

}

@incollection{dorigo2018introduction,
  title={An introduction to ant colony optimization},
  author={Dorigo, Marco and Socha, Krzysztof},
  booktitle={Handbook of approximation algorithms and metaheuristics},
  pages={395--408},
  year={2018},
 
}

@article{zheng2025towards,
  title={Towards lifelong learning of large language models: A survey},
  author={Zheng, Junhao and Qiu, Shengjie and Shi, Chengming and Ma, Qianli},
  journal={ACM Computing Surveys},
  volume={57},
  number={8},
  pages={1--35},
  year={2025},
}

@article{hu2024routerbench,
  title={Routerbench: A benchmark for multi-llm routing system},
  author={Hu, Qitian Jason and Bieker, Jacob and Li, Xiuyu and Jiang, Nan and Keigwin, Benjamin and Ranganath, Gaurav and Keutzer, Kurt and Upadhyay, Shriyash Kaustubh},
  journal={arXiv preprint arXiv:2403.12031},
  year={2024}
}

@article{li2025parallelized,
  title={Parallelized Planning-Acting for Efficient LLM-based Multi-Agent Systems},
  author={Li, Yaoru and Liu, Shunyu and Zheng, Tongya and Song, Mingli},
  journal={arXiv preprint arXiv:2503.03505},
  year={2025}
}

@article{yang2025docagent,
  title={DocAgent: A Multi-Agent System for Automated Code Documentation Generation},
  author={Yang, Dayu and Simoulin, Antoine and Qian, Xin and Liu, Xiaoyi and Cao, Yuwei and Teng, Zhaopu and Yang, Grey},
  journal={arXiv preprint arXiv:2504.08725},
  year={2025}
}

@article{yuan2024evoagent,
  title={Evoagent: Towards automatic multi-agent generation via evolutionary algorithms},
  author={Yuan, Siyu and Song, Kaitao and Chen, Jiangjie and Tan, Xu and Li, Dongsheng and Yang, Deqing},
  journal={arXiv preprint arXiv:2406.14228},
  year={2024}
}

@article{motwani2024malt,
  title={Malt: Improving reasoning with multi-agent llm training},
  author={Motwani, Sumeet Ramesh and Smith, Chandler and Das, Rocktim Jyoti and Rafailov, Rafael and Laptev, Ivan and Torr, Philip HS and Pizzati, Fabio and Clark, Ronald and de Witt, Christian Schroeder},
  journal={arXiv preprint arXiv:2412.01928},
  year={2024}
}

@article{jin2025comprehensive,
  title={A comprehensive survey on multi-agent cooperative decision-making: Scenarios, approaches, challenges and perspectives},
  author={Jin, Weiqiang and Du, Hongyang and Zhao, Biao and Tian, Xingwu and Shi, Bohang and Yang, Guang},
  journal={arXiv preprint arXiv:2503.13415},
  year={2025}
}

@article{wang2024cooperation,
  title={A cooperation and decision-making framework in dynamic confrontation for multi-agent systems},
  author={Wang, Lexing and Qiu, Tenghai and Pu, Zhiqiang and Yi, Jianqiang},
  journal={Computers and Electrical Engineering},
  volume={118},
  pages={109300},
  year={2024},
 
}

@article{chuang2025confident,
  title={Confident or seek stronger: Exploring uncertainty-based on-device llm routing from benchmarking to generalization},
  author={Chuang, Yu-Neng and Yu, Leisheng and Wang, Guanchu and Zhang, Lizhe and Liu, Zirui and Cai, Xuanting and Sui, Yang and Braverman, Vladimir and Hu, Xia},
  journal={arXiv preprint arXiv:2502.04428},
  year={2025}
}

@article{wang2024mixture,
  title={Mixture-of-agents enhances large language model capabilities},
  author={Wang, Junlin and Wang, Jue and Athiwaratkun, Ben and Zhang, Ce and Zou, James},
  journal={arXiv preprint arXiv:2406.04692},
  year={2024}
}

@article{marey2024explainability,
  title={Explainability, transparency and black box challenges of AI in radiology: Impact on patient care in cardiovascular radiology},
  author={Marey, Ahmed and Arjmand, Parisa and Alerab, Ameerh Dana Sabe and Eslami, Mohammad Javad and Saad, Abdelrahman M and Sanchez, Nicole and Umair, Muhammad},
  journal={Egyptian Journal of Radiology and Nuclear Medicine},
  volume={55},
  number={1},
  pages={183},
  year={2024},
}

@article{varangot2025doing,
  title={Doing More with Less--Implementing Routing Strategies in Large Language Model-Based Systems: An Extended Survey},
  author={Varangot-Reille, Clovis and Bouvard, Christophe and Gourru, Antoine and Ciancone, Mathieu and Schaeffer, Marion and Jacquenet, Fran{\c{c}}ois},
  journal={arXiv e-prints},
  pages={arXiv--2502},
  year={2025}
}

@article{zhang2024edgeshard,
  title={Edgeshard: Efficient llm inference via collaborative edge computing},
  author={Zhang, Mingjin and Shen, Xiaoming and Cao, Jiannong and Cui, Zeyang and Jiang, Shan},
  journal={IEEE Internet of Things Journal},
  year={2024},

}

@article{ding2024hybrid,
  title={Hybrid llm: Cost-efficient and quality-aware query routing},
  author={Ding, Dujian and Mallick, Ankur and Wang, Chi and Sim, Robert and Mukherjee, Subhabrata and Ruhle, Victor and Lakshmanan, Laks VS and Awadallah, Ahmed Hassan},
  journal={arXiv preprint arXiv:2404.14618},
  year={2024}
}

@article{cobbe2021training,
  title={Training verifiers to solve math word problems},
  author={Cobbe, Karl and Kosaraju, Vineet and Bavarian, Mohammad and Chen, Mark and Jun, Heewoo and Kaiser, Lukasz and Plappert, Matthias and Tworek, Jerry and Hilton, Jacob and Nakano, Reiichiro and others},
  journal={arXiv preprint arXiv:2110.14168},
  year={2021}
}

@article{hendrycks2024measuring,
  title={Measuring mathematical problem solving with the math dataset, 2021},
  author={Hendrycks, Dan and Burns, Collin and Kadavath, Saurav and Arora, Akul and Basart, Steven and Tang, Eric and Song, Dawn and Steinhardt, Jacob},
  journal={URL https://arxiv. org/abs/2103.03874},
  year={2024}
}

@article{hendrycks2020measuring,
  title={Measuring massive multitask language understanding},
  author={Hendrycks, Dan and Burns, Collin and Basart, Steven and Zou, Andy and Mazeika, Mantas and Song, Dawn and Steinhardt, Jacob},
  journal={arXiv preprint arXiv:2009.03300},
  year={2020}
}

@article{chen2021evaluating,
  title={Evaluating large language models trained on code},
  author={Chen, Mark and Tworek, Jerry and Jun, Heewoo and Yuan, Qiming and Pinto, Henrique Ponde De Oliveira and Kaplan, Jared and Edwards, Harri and Burda, Yuri and Joseph, Nicholas and Brockman, Greg and others},
  journal={arXiv preprint arXiv:2107.03374},
  year={2021}
}

@article{austin2021program,
  title={Program synthesis with large language models},
  author={Austin, Jacob and Odena, Augustus and Nye, Maxwell and Bosma, Maarten and Michalewski, Henryk and Dohan, David and Jiang, Ellen and Cai, Carrie and Terry, Michael and Le, Quoc and others},
  journal={arXiv preprint arXiv:2108.07732},
  year={2021}
}

@inproceedings{zhuge2024gptswarm,
  title={Gptswarm: Language agents as optimizable graphs},
  author={Zhuge, Mingchen and Wang, Wenyi and Kirsch, Louis and Faccio, Francesco and Khizbullin, Dmitrii and Schmidhuber, J{\"u}rgen},
  booktitle={Forty-first International Conference on Machine Learning},
  year={2024}
}

@article{qian2024scaling,
  title={Scaling large-language-model-based multi-agent collaboration},
  author={Qian, Chen and Xie, Zihao and Wang, Yifei and Liu, Wei and Dang, Yufan and Du, Zhuoyun and Chen, Weize and Yang, Cheng and Liu, Zhiyuan and Sun, Maosong},
  journal={arXiv preprint arXiv:2406.07155},
  year={2024}
}

@inproceedings{chen2023agentverse,
  title={Agentverse: Facilitating multi-agent collaboration and exploring emergent behaviors},
  author={Chen, Weize and Su, Yusheng and Zuo, Jingwei and Yang, Cheng and Yuan, Chenfei and Chan, Chi-Min and Yu, Heyang and Lu, Yaxi and Hung, Yi-Hsin and Qian, Chen and others},
  booktitle={The Twelfth International Conference on Learning Representations},
  year={2023}
}

@article{zhang2024aflow,
  title={Aflow: Automating agentic workflow generation},
  author={Zhang, Jiayi and Xiang, Jinyu and Yu, Zhaoyang and Teng, Fengwei and Chen, Xionghui and Chen, Jiaqi and Zhuge, Mingchen and Cheng, Xin and Hong, Sirui and Wang, Jinlin and others},
  journal={arXiv preprint arXiv:2410.10762},
  year={2024}
}

@article{yao2023tree,
  title={Tree of thoughts: Deliberate problem solving with large language models},
  author={Yao, Shunyu and Yu, Dian and Zhao, Jeffrey and Shafran, Izhak and Griffiths, Tom and Cao, Yuan and Narasimhan, Karthik},
  journal={Advances in neural information processing systems},
  volume={36},
  pages={11809--11822},
  year={2023}
}

@inproceedings{besta2024graph,
  title={Graph of thoughts: Solving elaborate problems with large language models},
  author={Besta, Maciej and Blach, Nils and Kubicek, Ales and Gerstenberger, Robert and Podstawski, Michal and Gianinazzi, Lukas and Gajda, Joanna and Lehmann, Tomasz and Niewiadomski, Hubert and Nyczyk, Piotr and others},
  booktitle={Proceedings of the AAAI conference on artificial intelligence},
  volume={38},
  number={16},
  pages={17682--17690},
  year={2024}
}

@article{zhang2025agentrouter,
  title={AgentRouter: A Knowledge-Graph-Guided LLM Router for Collaborative Multi-Agent Question Answering},
  author={Zhang, Zheyuan and Shi, Kaiwen and Yuan, Zhengqing and Wang, Zehong and Ma, Tianyi and Murugesan, Keerthiram and Galassi, Vincent and Zhang, Chuxu and Ye, Yanfang},
  journal={arXiv preprint arXiv:2510.05445},
  year={2025}
}

@article{liu2025rcr,
  title={Rcr-router: Efficient role-aware context routing for multi-agent llm systems with structured memory},
  author={Liu, Jun and Kong, Zhenglun and Yang, Changdi and Yang, Fan and Li, Tianqi and Dong, Peiyan and Nanjekye, Joannah and Tang, Hao and Yuan, Geng and Niu, Wei and others},
  journal={arXiv preprint arXiv:2508.04903},
  year={2025}
}

@article{zhao2026tcandon,
  title={TCAndon-Router: Adaptive Reasoning Router for Multi-Agent Collaboration},
  author={Zhao, Jiuzhou and Chen, Chunrong and Qiao, Chenqi and Zheng, Lebin and Han, Minqi and Zhang, Yanchi Liu Yongzhou Xu Xiaochuan Xu Min},
  journal={arXiv preprint arXiv:2601.04544},
  year={2026}
}

@article{lacavalla2024henry,
  title={HEnRY: A Multi-Agent System Framework for Multi-Domain Contexts},
  author={Lacavalla, Emmanuele and Yang, Shuyi and Crupi, Riccardo and Gonzalez, Joseph E},
  journal={arXiv preprint arXiv:2410.12720},
  year={2024}
}

@article{teng2025atom,
  title={Atom of thoughts for markov llm test-time scaling},
  author={Teng, Fengwei and Shi, Quan and Yu, Zhaoyang and Zhang, Jiayi and Luo, Yuyu and Wu, Chenglin and Guo, Zhijiang},
  journal={arXiv preprint arXiv:2502.12018},
  year={2025}
}

@article{cemri2025multi,
  title={Why do multi-agent llm systems fail?},
  author={Cemri, Mert and Pan, Melissa Z and Yang, Shuyi and Agrawal, Lakshya A and Chopra, Bhavya and Tiwari, Rishabh and Keutzer, Kurt and Parameswaran, Aditya and Klein, Dan and Ramchandran, Kannan and others},
  journal={arXiv preprint arXiv:2503.13657},
  year={2025}
}

@article{aratchige2025llms,
  title={Llms working in harmony: A survey on the technological aspects of building effective llm-based multi agent systems},
  author={Aratchige, RM and Ilmini, WMKS},
  journal={arXiv preprint arXiv:2504.01963},
  year={2025}
}

@article{chacon2025cooperative,
  title={Cooperative resilience in artificial intelligence multiagent systems},
  author={Chacon-Chamorro, Manuela and Giraldo, Luis Felipe and Quijano, Nicanor and Vargas-Panesso, Vicente and Gonz{\'a}lez, C{\'e}sar and Pinz{\'o}n, Juan Sebasti{\'a}n and Manrique, Rub{\'e}n and R{\'\i}os, Manuel and Fonseca, Yesid and G{\'o}mez-Barrera, Daniel and others},
  journal={IEEE Transactions on Artificial Intelligence},
  year={2025},
  publisher={IEEE}
}

@article{mosquera2025can,
  title={Can LLM-augmented autonomous agents cooperate? An evaluation of their cooperative capabilities through melting pot},
  author={Mosquera, Manuel and Pinzon, Juan Sebastian and Fonseca, Yesid and R{\'\i}os, Manuel and Quijano, Nicanor and Giraldo, Luis Felipe and Manrique, Ruben},
  journal={IEEE Transactions on Artificial Intelligence},
  year={2025},
  publisher={IEEE}
}

@article{hu2024scalable,
  title={Scalable learning for multiagent route planning: Adapting to diverse task scales},
  author={Hu, Guoqiang},
  journal={IEEE Transactions on Artificial Intelligence},
  volume={5},
  number={10},
  pages={4996--5011},
  year={2024},
  publisher={IEEE}
}

@article{dorigo2018ant,
  title={Ant colony optimization: overview and recent advances},
  author={Dorigo, Marco and St{\"u}tzle, Thomas},
  journal={Handbook of metaheuristics},
  pages={311--351},
  year={2018},
  publisher={Springer}
}

@article{li2025distributed,
  title={Distributed Reinforcement Learning Optimal Cluster Consensus Control for Takagi-Sugeno Fuzzy Multi-Agent Systems},
  author={Li, Hui and Ning, Jun and Tong, Shaocheng},
  journal={IEEE Transactions on Artificial Intelligence},
  year={2025},
  publisher={IEEE}
}

@article{adornetto2025generative,
  title={Generative agents in agent-based modeling: Overview, validation, and emerging challenges},
  author={Adornetto, Carlo and Mora, Adrian and Hu, Kai and Garcia, Leticia Izquierdo and Atchade-Adelomou, Parfait and Greco, Gianluigi and Pastor, Luis Alberto Alonso and Larson, Kent},
  journal={IEEE Transactions on Artificial Intelligence},
  year={2025},
  publisher={IEEE}
}

@inproceedings{zhanglearning,
  title={Learning Global Hypothesis Space for Enhancing Synergistic Reasoning Chain},
  author={Zhang, Jiaquan and Zhang, Chaoning and Chen, Shuxu and Wang, Xudong and Huang, Zhenzhen and Zheng, Pengcheng and Yuan, Shuai and Zheng, Sheng and Sun, Qigan and Zou, Jie and others},
  booktitle={The Fourteenth International Conference on Learning Representations}
}

@article{zhang2026text,
  title={Text summarization via global structure awareness},
  author={Zhang, Jiaquan and Zhang, Chaoning and Chen, Shuxu and Liu, Yibei and Li, Chenghao and Sun, Qigan and Yuan, Shuai and Puspitasari, Fachrina Dewi and Han, Dongshen and Wang, Guoqing and others},
  journal={arXiv preprint arXiv:2602.09821},
  year={2026}
}
\end{document}